\documentclass[sigconf]{acmart}

\usepackage{booktabs} 
\usepackage{amsmath}
\usepackage{amssymb}
\usepackage{tikz}
\usepackage{subfig}
\usetikzlibrary{bayesnet}

\usepackage{enumitem}

\copyrightyear{2018} 
\acmYear{2018} 
\setcopyright{rightsretained}
\acmConference[RecSys '18]{Twelfth ACM Conference on Recommender Systems}{October 2--7, 2018}{Vancouver, BC, Canada}
\acmPrice{15.00}
\acmDOI{10.1145/3240323.3240388}
\acmISBN{978-1-4503-5901-6/18/10}

\setlist[itemize]{leftmargin=*,noitemsep,nolistsep,topsep=0pt}

\begin{document}
\title{A Hierarchical Bayesian Model for Size Recommendation in Fashion}

\author{Romain Guigour\`{e}s}
\email{romain.guigoures@zalando.de}
\affiliation{%
  \institution{Zalando SE}
}
\author{Yuen King Ho}
\email{yuen.king.ho@zalando.de}
\affiliation{%
  \institution{Zalando SE}
}
\author{Evgenii Koriagin}
\email{evgenii.koriagin@zalando.de}
\affiliation{%
  \institution{Zalando SE}
}
\author{Abdul-Saboor Sheikh}
\email{saboor.sheikh@zalando.de}
\affiliation{%
  \institution{Zalando Research}
}
\author{Urs Bergmann}
\email{urs.bergmann@zalando.de}
\affiliation{%
  \institution{Zalando Research}
}
\author{Reza Shirvany}
\email{reza.shirvany@zalando.de}
\affiliation{%
  \institution{Zalando SE}
}

\begin{abstract}
We introduce a hierarchical Bayesian approach to tackle the challenging problem of size recommendation in e-commerce fashion. Our approach jointly models a size purchased by a customer, and its possible return event:  1. no return, 2. returned too small 3. returned too big.  Those events are drawn following a multinomial distribution parameterized on the joint probability of each event, built following a hierarchy combining priors. Such a model allows us to incorporate extended domain expertise and article characteristics as prior knowledge, which in turn makes it possible for the underlying parameters to emerge thanks to sufficient data. Experiments are presented on real (anonymized) data from millions of customers along with a detailed discussion on the efficiency of such an approach within a large scale production system.
\end{abstract}

\maketitle

\renewcommand{\shortauthors}{R. Guigour\`{e}s et al.}
\section{Introduction}

Fashion is a way to express identity, moods and opinions. Customers also tend to use fashion to either emphasize certain parts of their body or hide others. In that context, size and fit have been shown to be among factors influencing the most the overall satisfaction \cite{Pisut2017}. Online customers have to buy before trying their clothes on. The sensory feedback phase about how the article fits via touch and visual cues is then delayed. Because of these uncertainties, a lot of consumers are still reluctant to engage in the purchase process. 

To make matters worse, fashion articles including shoes and apparel have important sizing variations primarily due to: 1. different definitions of respective sizes from brands: the size systems used for specific categories are limited (e.g. S, M, L, etc.), however the sizes themselves represent different physical measurements from one brand to another; 2. different ways of converting a local size system to another: in Europe, garment sizes are not standardized and brands don't always use the same conversion logic.

A way to circumvent the confusion created by these variations is to use size tables which map physical body measurements to the article size system, requiring customers to have accurate measurements of their body. However, size tables themselves might suffer from a large variance, up to one inch within a single size. These differences stem from either different datasets used for size tables (e.g. German vs. UK population) or are due to vanity sizing, i.e deliberate size inconsistencies in brands targeting a specific focus group based on age, sportiness, etc. which represent major influences on the body measurements \cite{ujevic2005, shin2007,faust2014}.   
The combination of the above factors leaves the customers alone to face a highly challenging problem of determining the right size and fit during their purchase journey. In recent years, there has been a lot of interest in building recommender systems in fashion e-commerce with major focus on modeling style preferences based on customers past interactions, taste and affinities \cite{hu2015,arora2016,bracher2016}. However, few research work have been conducted to tackle the size recommendation problem. 

The recommendation of size and fit has been recently studied in \cite{abdulla2017} where sparsity in purchased data is mentioned as a major issue, especially considering that articles have a limited stock. To minimize that problem, the authors propose to represent articles as a combination of brand, usage, size, and fit. A neural network is then trained to learn a latent vector describing each article defined as the combination of features mentioned before. Customer vector representation is obtained by aggregating over purchased articles and, finally, a gradient boosted classifier predicts the fit of an article to a customer.

Following a different approach, the authors of \cite{sembium2017} propose a solution for determining if an article of a certain size would be fit, large, or small for a certain customer, using the purchase history. This is achieved by iteratively deducing the true sizes for customers and products, fitting a linear function based on the difference in sizes, and performing ordinal regression on the output of the function to get the loss. Extra features are simply included by addition to the linear function. To handle multiple persons behind a single account, hierarchical clustering is performed on each customer account before doing the above. An extension of that work has been very recently published proposing a Bayesian approach on a similar model \cite{Sembium2018}. Instead of learning the parameters in an iterative process, the updates are done with mean-field variational inference with Polya-Gamma augmentation. This method therefore naturally benefits from the nice advantages of Bayesian modeling - the uncertainty outputs, and the use of priors. 
In this paper, we present two approaches: a baseline algorithm, launched on shoes in 2016 and on garments in 2017, which consists in inferring the size that a customer intends to buy and, independently, the article's sizing characteristics ; and a hierarchical Bayesian approach which aims at jointly modeling the purchases of one or multiple sizes of an article along with their possible return events: 1. no return (article is kept), 2. returned too small 3. returned too big. 

In the size recommendation context, data sparsity is severe since it affects both articles and customers. To tackle this, we propose two design choices: a) building a hierarchy on top of parameters, exploiting prior knowledge on articles and customers; b) virtually treating the size as a continuous variable in the training phase.

\section{Methodology}

Customers experience in fashion e-commerce consists in selecting an article in a desired size, trying the article, forming an opinion on its size and returning or keeping it. To simulate the customers behavior towards sizing, we model the joint probability of a customer to pick a size and the resulting return status. Return status is described by three possible events: the customer keeps the article, the customers returns the article because it's too small and the customer returns the article because it is too big. We ignore the cases where the customer returns the article for any other reason. 

\subsection{Notation}

Let us denote $\mathcal{C}$ the set of customers and $\mathcal{A}$ the set of articles. The size $\mathcal{S}_i$ is a continuous random variable. The variable $\mathcal{R}$ indicates the return status described above. Orders $\mathcal{O}$ are defined by a customer, an article, the purchased size and the return status. Both approaches introduced in the paper model the joint probability $p(\mathcal{S}_o, \mathcal{R}_o \mid \mathcal{C}_o, \mathcal{A}_o)$ as detailed in the following.
\subsection{Baseline Model}
\label{sec:baseline}
The baseline model makes a simplifying assumption that the size the customer chose and the return status are two independent events. Thus, the joint probability is defined as the product of the probability over sizes and the probability of return status: 

\begin{equation}
p(\mathcal{S}, \mathcal{R} \mid \mathcal{C}, \mathcal{A}) = p(\mathcal{S}  \mid \mathcal{C}, \mathcal{A}) p(\mathcal{R} \mid\mathcal{C}, \mathcal{A})
\end{equation}

\paragraph{Probability over sizes} 

We assume that multiple persons can use a single account. The probability distribution over sizes is obtained by Gaussian Kernel Density Estimation. To avoid getting degenerate distributions when customers always purchase the same size, we set a lower limit on the variances. Let denote $\mathcal{O}_j$ the set of $n_j$ orders and $S_j = \{s_i, i=1..n_j\}$ the set of $n_j$ sizes purchased by the customer $c_j$. The related probability density function is defined as: 

\begin{equation}
p(s \mid c_j) = \dfrac{1}{n_jh_j} \displaystyle\sum_{i=1}^{n_j} \phi\left(\dfrac{s-s_i}{h_j}\right)
\end{equation}

where $\phi$ is the normal density function and $h_j$ is the bandwidth parameter for that specific customer $c_j$. The latter is obtained by minimizing the mean integrated squared error \cite{Silverman1986}. 

\paragraph{Probability over return status}

Customers have different return behaviors. However, impact of customers on returns is assumed negligible compared to potential sizing issues of the article. The probability of each return status is  consequently marginalized over customers: $p(\mathcal{R} \mid\mathcal{C}, \mathcal{A}) = p(\mathcal{R} \mid \mathcal{A})$. %
The probability over return status is the empirical distribution over the three possible events: article is kept, too big or too small. The case of one of the events being not observed in the training data may lead to a null probability in validation. To avoid that problem, we add one to the counts of each event. For an article $a_i$, sold $n_i$ times, the probability of a return event $r$, observed $n_{i,r}$ times in the data, is defined as:
\begin{equation}
p(r \mid a_i) = \dfrac{n_{i,r} + 1}{n_i + 3}
\end{equation}

Though, this method seems inelegant, it has a Bayesian grounding. Indeed, this is equivalent to taking the maximum a posteriori of a categorical distribution with a Dirichlet conjugate prior, which concentration parameter is equal to one, i.e the uniform distribution. 
In case of a cold start, i.e if a customer (resp. an article) is new, the marginal distribution over sizes of all the customers (resp. over return status of all articles) is used.

\subsection{Hierarchical Bayesian Model}
\label{sec:BHM}
The baseline described above has a risk of specious parameter estimation and overfitting. Bayesian approaches conversely aim at providing a probability of the estimated parameters given a set of observed data, supporting the decision process when offering a recommendation to the customer. Therefore, using a Bayesian approach, we aim at modeling the joint probability of a size to be purchased and a return status to be observed without the simplifying hypothesis from the base approach. For each pair of customer and article, orders $\mathcal{O}$ are drawn following a categorical distribution.

\begin{equation}
\mathcal{O} \sim Cat(p(\mathcal{S}, \mathcal{R} \mid \mathcal{C}, \mathcal{A}))
\end{equation}

Contrary to the baseline, both $\mathcal{S}$ and $\mathcal{R}$ are not assumed independent, instead the joint probability is factorized as: 

\begin{equation}
p(\mathcal{S}, \mathcal{R} \mid \mathcal{C}, \mathcal{A}) = p(\mathcal{S} \mid \mathcal{R}, \mathcal{C}, \mathcal{A}) \times p(\mathcal{R} \mid \mathcal{C}, \mathcal{A})
\label{eq:factorization}
\end{equation}

It is worth noting that methods described in \cite{Sembium2018,abdulla2017} aim at modeling $p(\mathcal{R} \mid \mathcal{S}, \mathcal{C}, \mathcal{A})$. Doing so requires discretizing the continuous variable $\mathcal{S}$ leading to an increase in the number of parameters to be inferred and making the model more susceptible to the sparsity of the data. The factorization we chose in Equation \ref{eq:factorization} allows us to model $p(\mathcal{S} \mid \mathcal{R}, \mathcal{C}, \mathcal{A})$ as a continuous distribution. This enables us to learn a smaller set of parameters specifying the distribution over all sizes, which helps to alleviate part of the sparsity problem. 

\paragraph{Probability over return status}

For the same reasons as explained in the Section \ref{sec:baseline}, the probability of return status is marginalized over customers: $p(\mathcal{R} \mid \mathcal{C}, \mathcal{A}) = p(\mathcal{R} \mid \mathcal{A})$. Returns are assumed independent from one another, allowing us to model them using a categorical distribution. As the number of purchases might be low for some articles, a Dirichlet prior is used. The concentration parameter of the prior is based on the counts at the brand level and at the category level (e.g. dresses, t-shirts, sneakers, etc.). Let $n_K$, $n_S$, and $n_B$ denote the counts of kept articles, returned articles for being too small and too big at the article level respectively. In a similar way, $m_K$, $m_S$ and $m_B$ indicates the counts at the brand level and $m'_K$, $m'_S$ and $m’_B$, at the category level.

\begin{equation}
p(\mathcal{R} \mid \mathcal{A}) \sim Dirichlet(\alpha)
\end{equation}
with $\alpha = w \cdot [m_K, m_S, m_B] + w’ \cdot [m'_K, m'_S, m'_B]$. The weights $w$ and $w'$ are learned under the assumption that they follow Beta distribution with a low first shape parameter and the second shape parameter equal to 1, in order to favor low weight values. 
\begin{equation}
w \sim Beta(0.5, 1) \mbox{ and } w' \sim Beta(0.1, 1)
\end{equation}
Since the Dirichlet prior is the conjugate of the categorical distribution, the posterior probability can be analytically computed, easing the inference of the parameters of the model. 
\begin{equation}
P(\mathcal{R}=r \mid \mathcal{A}, \mathcal{O} ) = \dfrac{n_r + \alpha_r}{\displaystyle\sum_{\substack{i \in \\ \{K,S,B\}}} n_i + \displaystyle\sum_{\substack{i \in \\ \{K,S,B\}}}\alpha_i}
\end{equation}

\paragraph{Probability over sizes conditionally on the return status}

For a given customer and article, the probability distribution of the customer buying a size is a mixture of Gaussians. Since the number of users of an account is unknown, we decide to use an infinite mixture model. However, we assume that the number of distinct persons using a single account is low. That's why we opt for a Dirichlet process with a truncation level fixed to four. In order to ease the inference, we use a truncated stick-breaking process \cite{sethuraman1994}. 
\begin{equation}
\begin{split}
&p(s \mid r, c, a) = \displaystyle\sum_{i=1}^{4} \pi_i \phi(s \mid \mu_i, \sigma_i^{2}) \\
&\pi_i = b_i \displaystyle\prod_{j=1}^{i-1} (1-b_j) \\
&b_i \sim Beta(1, \alpha) \mbox{ for }  i=1..3 \mbox{ and } b_4 = 1
\end{split}
\label{eq:mix_gaussian}
\end{equation}
The parameter $\pi$ is the mixing proportion, that can be interpreted as the probability of person $i$ using the account of the customer $c$. The shape parameter $\alpha$ of the $Beta$ distribution is acting as a concentration parameter of the Dirichlet process: the case $\alpha = 1$ is equivalent to the uniform distribution, thus favoring the scenario of multiple persons sharing a single account; conversely, the case $\alpha \rightarrow 0$ produces high density around 1 which favors the scenario of a single person placing all the orders. In the context of size recommendation, $\alpha$ is fixed at $0.5$.

\begin{figure}[!h]
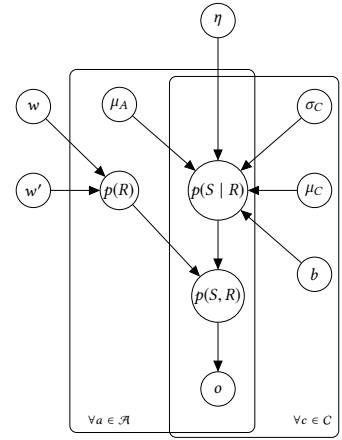

\scalebox{.65}{\tikz{ %
	\node[latent] (o) {$o$} ; %
    \node[latent, above=of o] (pjoint) {$p(S,R)$} ; %
    \node[latent, above =of pjoint] (psr) {$p(S\mid R)$} ; 
            \node[latent, right =of psr] (muC) {$\mu_C$} ; %
             \node[latent, below =of muC] (b) {$b$} ; %
            \node[latent, above =of muC] (sigmaC) {$\sigma_C$} ; %
    \node[latent,  left=of psr] (pr) {$p(R)$} ; %
    \node[latent, above =of pr] (muA) {$\mu_A$} ; %
    \node[latent, above=of muA, draw=white] (hidden) {} ; %
    \node[latent, right=of hidden, xshift=0.3cm] (eta) {$\eta$} ; %
    \node[latent, left =of pr] (wprime) {$w'$} ; %
    \node[latent, above=of wprime] (w) {$w$} ; %
    
    \plate[inner sep=0.25cm, xshift=-0.12cm, yshift=0.1cm] {plate1} {(pr) (pjoint) (psr) (o) (muA)} {$\forall a \in \mathcal{A} \phantom{xxxxxxxxxxxxxxx}$}; %
    \plate[inner sep=0.25cm, xshift=-0.12cm, yshift=0.0cm] {plate2} {(psr) (pjoint) (o) (muC) (b) (sigmaC)} {$\forall c \in \mathcal{C}$}; %
    
    \edge {w} {pr} ; %
    \edge {wprime} {pr} ; %
     \edge {pr} {pjoint} ; %
     \edge {muA} {psr} ; 
     \edge {muC} {psr} ; 
     \edge {sigmaC} {psr} ; 
     \edge {b} {psr} ; 
     \edge {eta} {psr} ; 
    \edge {psr} {pjoint} ; %
     \edge {pjoint} {o} ; %

 }}
 \caption{Graphical model of the Bayesian approach}
\label{fig:bayesian_graphe}
 \end{figure}

The function $\phi$ in equation \ref{eq:mix_gaussian} is the normal probability density function over sizes for the person $i$ using the account of the customer $c$, buying an article $a$, resulting in a return status $r$. The parameter $\mu$ is a combination of three parameters $\mu = \mu_C + \mu_A + \eta_R$:
\begin{itemize}
\item the average size of a person $\mu_C$,\begin{equation}
\mu_C \sim \mathcal{N}(\mu_0, \sigma_0^2)
\end{equation}
where hyperparameters $\mu_0$ and $\sigma_0^2$ depend on the category, the gender of the article and the size system;
\item the average offset of the article $\mu_A$,
\begin{equation}
\mu_A \sim \mathcal{N}(0, 1) 
\end{equation}
where assumption is made that most articles have an accurate size, i.e. an offset of zero;
\item a shifting parameter $\eta = \{\eta_K, \eta_S, \eta_B\}$ for each return status: resp. article is kept, returned too small and returned too big,
\begin{equation}
\eta_K = 0  \mbox{ ; } \eta_S \sim \mathcal{N}(-1, 1) \mbox{ ; } \eta_B \sim \mathcal{N}(1, 1)
\end{equation}
shifting parameter is fixed at 0 for the case the article is kept, while it's sampled using a Gaussian distribution centered to 1 (resp. -1) when the customer has returned the article because it is too big (resp. too small). 
\end{itemize} 
We assume the parameter $\sigma$ in equation \ref{eq:mix_gaussian} depends on the customer only. It is sampled following an Inverse Gamma distribution, with the shape parameter $\gamma_1$ and the scale parameter $\gamma_2$.
\begin{equation}
\sigma_C ^2 \sim \Gamma^{-1}(\gamma_1,\gamma_2) 
\end{equation}
We fix the parameters of the distribution to $\gamma_1 = 1$ and $\gamma_2 = 2$, so that the mode of the Inverse Gamma distribution is equal to 1. Figure \ref{fig:bayesian_graphe} represents the graphical model of the approach.
\paragraph{Inference} 
Monte-Carlo Markov Chains are popular sampling methods for Bayesian inference. But those approaches are often slow to converge. Variational inference methods run faster than sampling based methods but also introduce an approximation bias that may lead to a bad estimation of the parameters. However, those approaches have demonstrated reasonable performances on Dirichlet processes \cite{Blei2006} and are well suited for problems involving large amount of data. The inference is consequently done using mean-field approximation.

\subsection{Providing a Size Recommendation}

The set of sizes of an article is a finite set and as a consequence the probability density function needs to be discretized. For a customer $c$ and an article $a$ with a set of $k$ sizes $\mathcal{S} = \{s_i, i=1..k\}$, the probability over sizes is discretized as follows:
\begin{equation}
p(s, r) = p(r) \dfrac{\displaystyle\int_{s-\frac{1}{2}\epsilon}^{s+\frac{1}{2}\epsilon} f(x) dx}{\displaystyle\int_{s_1-\frac{1}{2}\epsilon}^{s_k+\frac{1}{2}\epsilon} f(x) dx}
\end{equation}
where $f$ is the probability over sizes, marginal for the baseline and conditionally to the return status for the Bayesian model ; $\epsilon$ is equal to the step between two sizes.
To provide a size recommendation, we choose the size having the highest probability to be kept by the customer.

\section{Experiments}
For this experiment, anonymized purchase data is collected for adult shoes. The data consists of 14.5 million purchases, 3 million distinct customers and 73,000 distinct articles. As the data has a strong temporal component, cross validation is performed under the following conditions \cite{arlot2010}: 1. validation data occurs later than training data, 2. a period of three weeks - corresponding to the time to collect most returns from customers - is ignored between train set and validation set and 3. validation sets must not overlap. Both models are trained and cross-validated on the same data. We reported the average logarithm of the joint probability $p(s,r)$ over all observations in the validation set in Table \ref{tab:likelihood}. Higher numbers show better performances of the model. 

\begin{table}[!h]
  \caption{Average log joint probability}
  \vspace{-0.3cm}
  \begin{tabular}{ | l | l | l |}
    \hline
     & Baseline & Bayesian \\ \hline
    incl. unknown customers & -2.85 (± 0.15) & -2.35 (± 0.11)\\ \hline
    excl. unknown customers & -3.32 (± 0.26)& -1.83 (± 0.19)\\
    \hline
  \end{tabular}
\label{tab:likelihood}
\end{table}

Table \ref{tab:likelihood} compares likelihood for both approaches, including and excluding customers not observed in the training data. Bayesian approach shows better results in both cases, where results are the best when excluding unknown customers. Conversely, the baseline performs better when unknown customers are included. This is mainly due to the fact that the baseline overfits by putting high probability density on the events observed in the training set, and very low density on unseen events. 

\begin{figure}[!ht]
\centering
\subfloat{\includegraphics[width=0.47\textwidth]{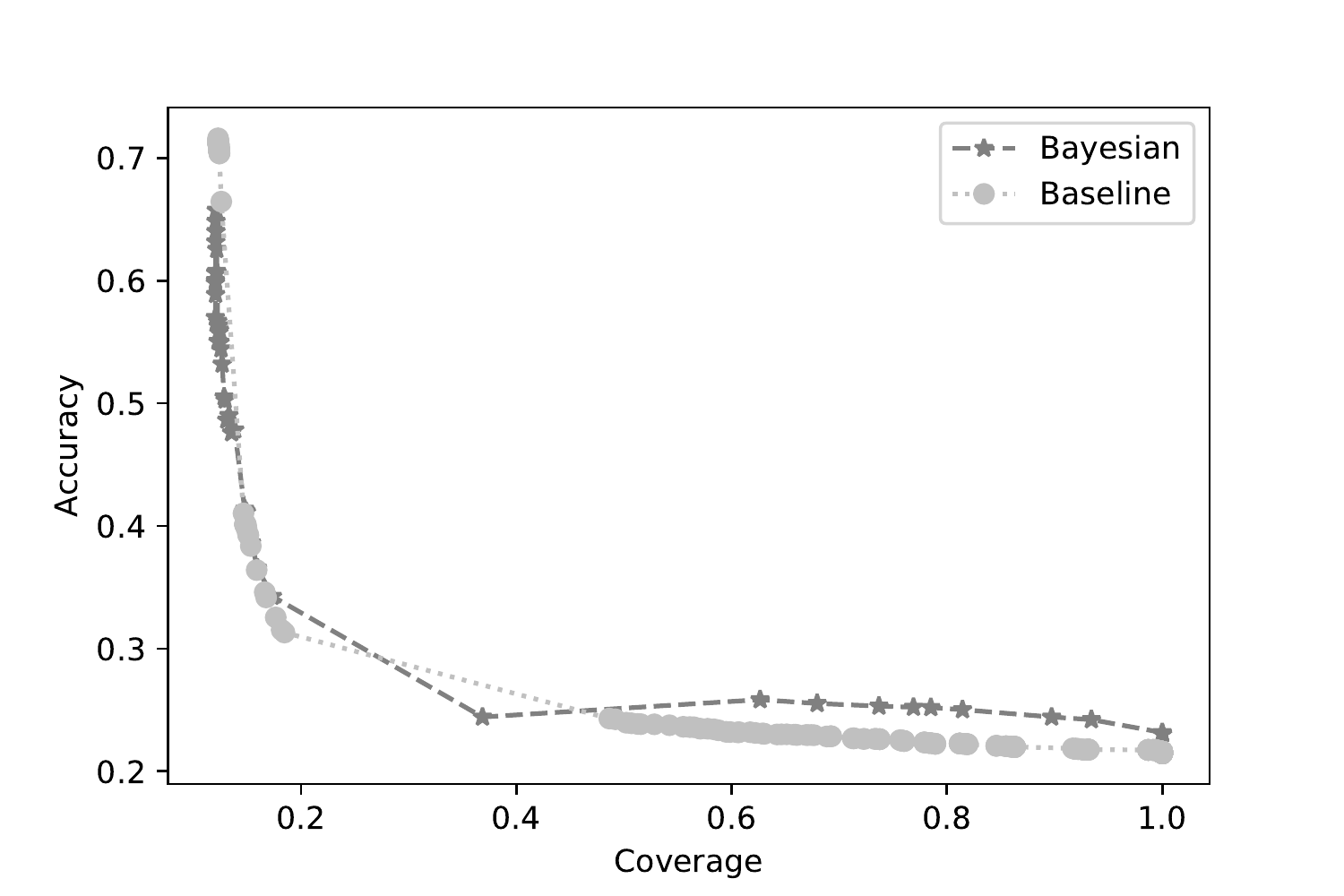}}
\quad
\subfloat{\includegraphics[width=0.47\textwidth]{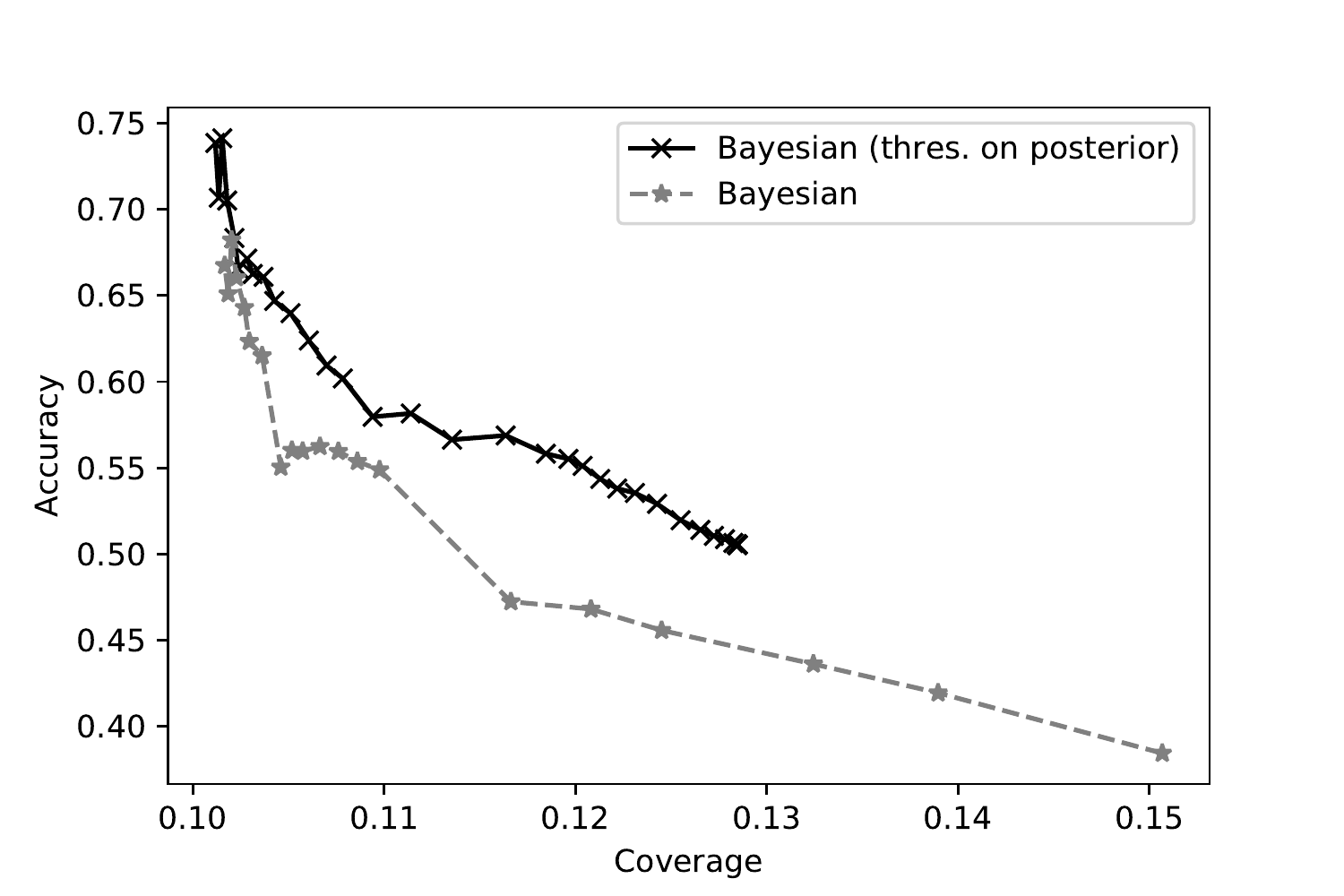}}
\caption{Accuracy versus coverage. (Top) Baseline and Bayesian model. (Bottom) Bayesian model with and without threshold on the posterior probability of the parameters.}
\label{fig:cov_vs_acc}
\end{figure}

In the context of size recommendation, two indicators play a key role in the decision process: coverage and accuracy. The coverage is the percentage of purchases for which the algorithm is confident making a decision. The accuracy is the number of correctly predicted sizes and return status, over all predictions. Figure \ref{fig:cov_vs_acc} (Top) shows the accuracy versus the coverage for several values of a sliding threshold on the joint probability for both models. On Figure \ref{fig:cov_vs_acc} (Bottom), the results are presented for the Bayesian model where we also include a threshold on the posterior probability of parameters.

Figure \ref{fig:cov_vs_acc} (Top) shows that accuracy decreases as coverage increases, when changing the threshold on the joint probability. The performances of the Baseline and the Bayesian model are similar, when the decision is based on the value of the joint probability. In Figure \ref{fig:cov_vs_acc} (Bottom), by putting a threshold on the posterior probability of parameters, we prevent the model from recommending article sizes to customers, where the parameters are poorly estimated. Performances are slightly better, however only 13\% of the purchases can be covered. 

It is worth mentioning that both approaches need to filter out a lot of purchases before starting to show reasonable accuracy levels. From the computational complexity point of view, inference of the Bayesian approach is more costly compared to the baseline model. Results are encouraging and demonstrate the complexity of the size recommender topic, motivating a deeper research work in the field. 
\section{Conclusion}
A hierarchical Bayesian approach was proposed to tackle the challenging problem of size recommendation in e-commerce fashion. The size purchased by a customer and its possible return events were jointly modeled thanks to a Bayesian approach. Experimental results were presented on real (anonymized) data from millions of customers along with a detailed discussion and comparison with a baseline approach with simplified hypothesis. It was shown that the Bayesian approach outperforms the baseline approach while providing better theoretical framework for gaining deeper understanding of the predictions. Future work consists in exploring different approaches to learn the joint probability, making use of additional article related data from fashion industry, along with deeper dives in the segmentation of customers and articles.
\bibliographystyle{unsrt} 
\bibliography{bibliography.bib}

\end{document}